\documentclass{article}
\usepackage{nips11submit_e,times,amsmath,caption,amsfonts,amssymb,cite,graphicx,caption,enumerate,stmaryrd}
\nipsfinalcopy 

\newcommand{\eod}{{${}$\\}}

\newcommand{\X}{{\mathcal{X}}}
\newcommand{\Y}{{\mathcal{Y}}}

\newcommand{\risk}{{\mathbf R}}

\newcommand{\slabel}{{\mathcal L}}
\newcommand{\data}{{\mathcal D}}

\newcommand{\bE}{{\mathbb E}}
\newcommand{\bI}{{\mathbb I}}

\newcommand{\bR}{{\mathbb R}}

\newcommand{\cA}{{\mathcal A}}

\newcommand{\cS}{{\mathcal S}}
\newcommand{\cV}{{\mathcal V}}
\newcommand{\cF}{{\mathcal F}}
\newcommand{\radem}{{\mathcal R}}

\newcommand{\fm}{{\mathfrak m}}

\newcommand{\errmap}{\risk}

\newtheorem{thm}{Theorem}
\newtheorem{prop}[thm]{Proposition}

\newtheorem{cor}[thm]{Corollary}

\newtheorem{defn}[thm]{Definition}
\newtheorem{rem}[thm]{Remark}

\begin{document}

\title{Falsification and future performance}
\author{
David Balduzzi\\
MPI for Intelligent Systems, T{\"u}bingen, Germany. \\
\texttt{david.balduzzi@tuebingen.mpg.de}
}

\maketitle
\begin{abstract}
	We information-theoretically reformulate two measures of capacity from statistical learning theory: empirical VC-entropy and empirical Rademacher complexity. We show these capacity measures count the number of hypotheses about a dataset that a learning algorithm \emph{falsifies} when it finds the classifier in its repertoire minimizing empirical risk. It then follows from that the future performance of predictors on \emph{unseen} data is controlled in part by how many hypotheses the learner falsifies. As a corollary we show that empirical VC-entropy quantifies the message length of the true hypothesis in the optimal code of a particular probability distribution, the so-called actual repertoire.
\end{abstract}

\section{Introduction}

This note relates the number of hypotheses falsified by a learning algorithm to the expected future performance of the predictor it outputs. It does so by reformulating two basic results from statistical learning theory information-theoretically.

Suppose we wish to predict an unknown physical process $\sigma^*:\X\rightarrow \Y$ occurring in nature after observing its outputs $(y_1,\ldots, y_l)$ on sample $\data=(x_1,\ldots,x_l)$ of its inputs, where inputs arise according to unknown distribution $P$. One method is to take a repertoire $\cF$ of functions from $\X\rightarrow \Y$ and choose the predictor $\hat{f}\in\cF$ that best approximates $\sigma^*$ on the observed data. How confident can we be in $\hat{f}$'s future performance on unseen data?

Statistical learning theory provides bounds on $\hat{f}$'s expected future performance by quantifying a tradeoff implicit in the choice of repertoire $\cF$. At first glance, the bigger the repertoire the better since the best approximation to $\sigma^*$ in $\cF$ can only improve as more more functions are added to $\cF$. However, increasing $\cF$, and improving the approximation on observed data, can \emph{reduce} future performance due to overfitting. As a result, the bounds depend on both the accuracy with which $\hat{f}$ approximates $\sigma^*$ on the observed data and the capacity of repertoire $\cF$, see Theorems~\ref{t:vc} and \ref{t:rademacher}.

We wish to connect statistical learning theory with Popper's ideas about falsification. Popper argued that no amount of positive evidence confirms a theory \cite{popper:59}. Rather, theories should be judged on the basis of how many hypotheses they falsify. A theory is \emph{falsifiable} if there are possible hypotheses about the world (i.e. data) that are not consistent with the theory. A bold theory falsifies (disagrees with) many potential hypotheses about observed data. Testing a bold theory, by checking that the hypotheses it disagrees with are in fact false, provides  corroborating evidence. If a theory has been thoroughly tested then (perhaps) we can have confidence in its predictions. Popper's criticism of positive confirmation was devastating. However, and hence the ``perhaps'', he failed to provide a rationale for trusting the predictions of severely tested theories.

To understand how falsifying hypotheses affects future performance we reformulate learning as a kind of \emph{measurement}. Before doing so, we need to describe precisely what we mean by measurement.

Given physical system $X$ with state space $S(X)$, a classical measurement is a function $f:S(X)\rightarrow \bR$. For example a thermometer $f$ maps configurations (positions and momenta) of particles in the atmosphere to real numbers. When the thermometer outputs $15^\circ C$ it generates information by specifying that atmospheric particles were in a configuration in $f^{-1}(15)\subset S(X)$. The information generated by the thermometer is a brute physical fact depending on how the thermometer is built and its output. We quantify the information, see \S\ref{s:meas}, by comparing the size of the total configuration space $S(X)$ with the size of the pre-image $f^{-1}(15)$. The smaller the pre-image, the more informative the measurement, see \S\ref{s:meas} for details.

More generally, any (classical) physical process $f:\X\rightarrow \Y$ can be thought of as performing measurements by taking inputs in $\X$ to outputs in $\Y$. Section~\S\ref{s:lim} introduces an important example, the \emph{min-risk} $\errmap_{\cF,\data}:\Sigma(\X,\Y)\rightarrow \bR$, which outputs the minimum value of the empirical risk over repertoire $\cF$ on a hypothesis space $\Sigma(\X,\Y)$. Finding the min-risk is a necessary step in finding the best approximation $\hat{f}$ to $\sigma^*$ in $\cF$. Since computing the min-risk requires actually implementing it as a physical process somehow or other, the measurements it performs and the effective information it generates are brute physical facts, no different in kind than the information generated by a thermometer.

It turns out that the min-risk categorizes hypotheses in $\Sigma$ according to how well they are approximated by predictors in repertoire $\cF$. Proposition~\ref{t:ei-vc} shows that the effective information generated by the min-risk is (essentially) the empirical VC-entropy. Moreover, the effective information generated by the min-risk ``counts'' the number of hypotheses about $\data$ that $\cF$ falsifies, see Eq.~\eqref{e:falsify}.  As a consequence, Corollary~\ref{t:ei-vcb}, we obtain that the future performance of predictor $\hat{f}$ is controlled by {(\rm i)} how well $\hat{f}$ fits the observed data; {(\rm ii)} how many hypotheses about the data the min-risk rules out and {(\rm iii)} a confidence term.

It follows that, assuming the assumptions of the theorems below hold, bounds on future performance are brute physical facts resulting from the act of minimizing empirical risk, and so falsifying potential hypotheses, on observed data. 

A consequence of our results, Corollary \ref{t:mml}, is that empirical VC-entropy is essentially the minimal length of the true hypothesis under the optimal code for the actual repertoire (a distribution depending on the min-risk). This suggests there may be interesting connections between VC-theory and the minimum message length (MML) approach to induction proposed by Wallace and Boulton \cite{wallace:68, wallace:05}.

Finally, section \S\ref{s:rrad} reformulates empirical Rademacher complexity via falsification. Here we build on Solomonoff's probability distribution introduced in \cite{solomonoff:64}. In short, we take Solomonoff's definition and substitute the \emph{min-risk} in place of the universal Turing machine, thereby obtaining what we refer to as the Rademacher distribution -- a \emph{non-universal} analog of Solomonoff's distribution. Rademacher complexity is then computed using the expectation of the min-risk over the Rademacher distribution, see Proposition \ref{t:mr-rad}. 

The min-risk thus provides a bridge that not only connects VC-theory to a computable analog of Solomonoff's seminal distribution, but also sheds light on how falsification provides guarantees on future performance.

\textbf{Related work.}
The connection between Popper's ideas on falsifiability and statistical learning theory was pointed out in \cite{vapnik:98, corfield:09, harman:07}. However, these works focus on VC-dimension, which does not relate to falsification as directly as VC-entropy and Rademacher complexity which we consider here. Further, VC-entropy is a more fundamental concept in statistical learning theory than VC-dimension since VC-dimension is defined in terms of the limit behavior of the growth function, which is an upper bound on VC-entropy \cite{vapnik:98}. For more details on the link between MML and algorithmic probability, see \cite{wallace:99}.

\textbf{Acknowledgements.}
I thank David Dowe and Samory Kpotufe for useful comments on an earlier version of this paper.

\section{Measurement}
\label{s:meas}

We consider a toy universe containing probabilistic mechanisms (input/output devices) of the following form
\begin{defn}
	Given finite sets $\X$ and $\Y$, a \textbf{mechanism} is a Markov matrix $\fm$ defined by conditional probability distribution  $p_\fm(y|x)$. 
\end{defn}
Mechanisms generate information about their inputs by assigning them to outputs \cite{bt:08, bt:09}. 
\begin{defn}
	The \textbf{actual repertoire} (or \textbf{measurement}) specified by $\fm$ outputting $y$ is the probability distribution
	\begin{equation*}
		p_\fm(x|y):=\frac{p_\fm(y|x)}{p(y)}\cdot p_{unif}(x),
	\end{equation*}
	where $p_{unif}(x)=\frac{1}{|\X|}$ is the uniform distribution. The \textbf{effective information} generated by the measurement is
	\begin{equation*}
		ei(\fm,y):=H\Big[p_\fm(X|y)\Big\|p_{unif}(X)\Big],
	\end{equation*}
	where $H[p\|q]=\sum_i p_i\log_2\frac{p_i}{q_i}$ is Kullback-Leibler divergence.
\end{defn}

The Kullback-Leibler divergence $H[p\|q]$ can be interpreted informally as the number of Y/N questions needed to get from distribution $q$ to distribution $p$. However, as pointed out in \cite{dowe:10}, Kullback-Leibler divergence is invariant with respect to the ``framing of the problem'' -- the ordering and structure of the questions -- suggesting it is a suitable measure of information-theoretic  ``effort''.

The definition of measurement is motivated by the special case where $p_\fm$ assigns probabilities that are either 0 or 1; in other words, when it corresponds to a set-valued function $f:\X\rightarrow \Y$. The measurement performed by $f$ is
\begin{equation*}
	p_f(x|y) = \left\{\begin{matrix}
		\frac{1}{|f^{-1}(y)|} & \mbox{ if }f(x)=y\\
		0 & \mbox{ else,}
	\end{matrix}\right.
\end{equation*}
where $|\cdot|$ denotes cardinality. The support of $p_f(X|y)$ is the preimage $f^{-1}(y)\subset \X$. All elements of the support are assigned equal probability -- they are treated as an undifferentiated list. The measurement $p_\fm(X|y)$ therefore generalizes the notion of preimage to the probabilistic setting. 

The effective information generated by $f$ outputting $y$ is $ei(f,y) = \log_2\frac{|\X|}{|f^{-1}(y)|}$:
\begin{equation}
	\begin{matrix}
		ei(f,y) &= & \log_2|\X|& -& \log_2|f^{-1}(y)|\\
		&=& \Big(\mbox{no. potential inputs}\Big) & - & \Big(\mbox{no. inputs in pre-image}\Big) \\
		&=& \Big(\mbox{no. inputs ruled out}\Big),
	\end{matrix}	
	\label{e:det-ei}
\end{equation}
where inputs are counted in bits (after logarithming). Effective information is maximal ($\log_2|\X|$ bits) when a single input leads to $y$, and is minimal (0 bits) when \emph{all} inputs lead to $y$. In the first case, observing $f$ output $y$ tells us exactly what the input was, and in the latter case, it tells us nothing at all.

\begin{figure}[thpb]
	\centering
	\includegraphics[scale=.7]{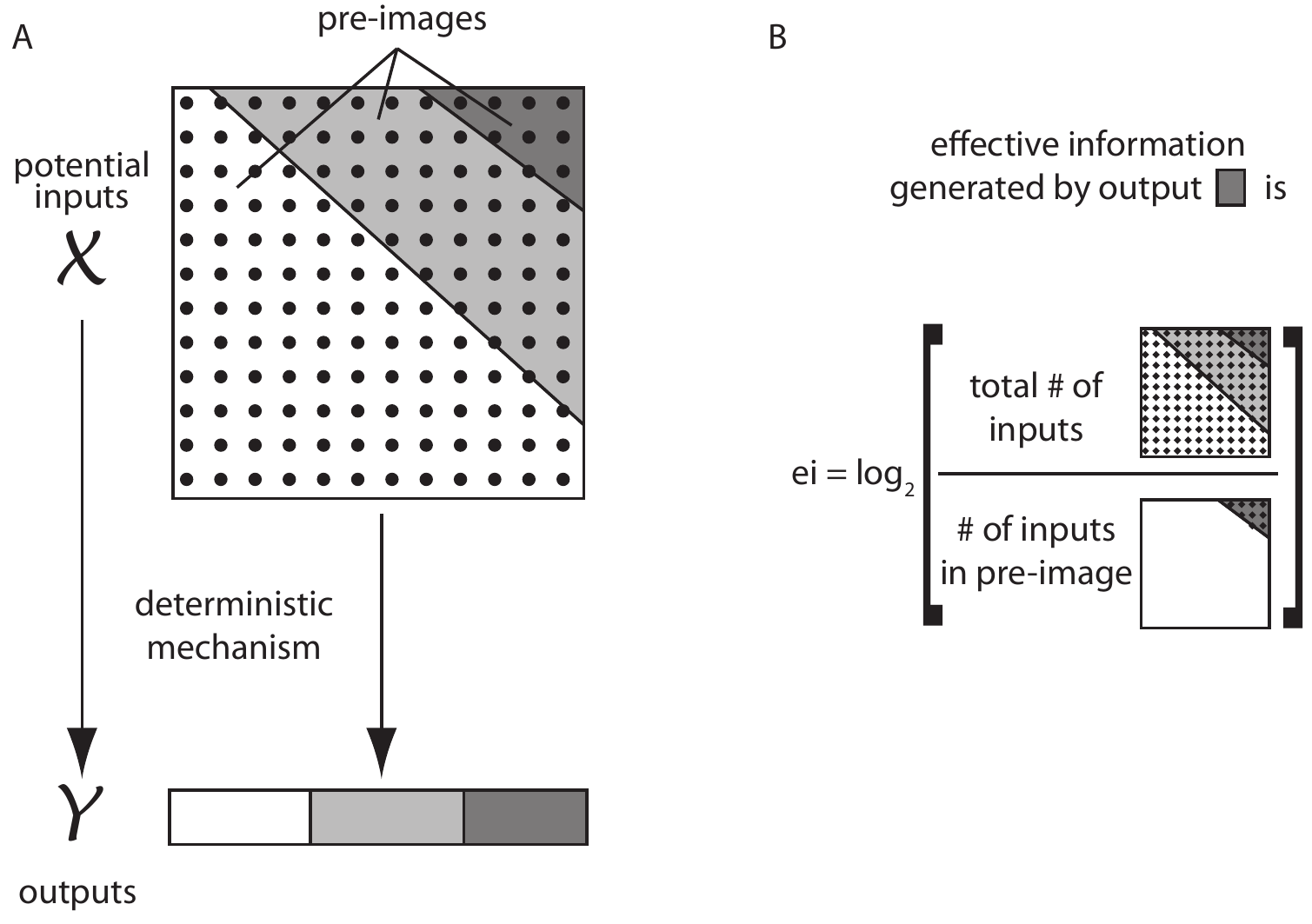}
	\caption{\textbf{The effective information generated by measurements.}
	\footnotesize{
	(A) A deterministic device can receive 144 inputs and produce 3 outputs. (B): Each input is implicitly assigned to a category (shaded areas). The information generated by the dark gray output is $\log_2 144-\log_2 9= 4$ bits.
	}}
	\label{f:meas}
\end{figure}

\subsection{Semantics}
\label{s:semantics}

Next we consider two approaches to characterizing the meaning of measurements. The first relates to possible world semantics \cite{lewis:86}. Here, the meaning of a sentence is given by the set of possible worlds in which it is true. Meaning is thus determined by considering all counterfactuals. For example, the meaning of ``That car is 10 years old'' is the set of possible worlds where the speaker is pointing to a car manufactured 10 years previously. Since the set of contains cars of many different colors, we see that color is irrelevant to the meaning of the sentence. 

More precisely, the meaning of sentence $\cS$ is a map from possible worlds $W$ to truth values $v_\cS:W\rightarrow\{0,1\}$. Equivalently, the meaning of a sentence is 
\begin{equation}
	\begin{matrix}
		W & \supset &  v_\cS^{-1}(1)\\
		\Big(\mbox{possible worlds}\Big) & \supset 
		& \Big(\mbox{worlds where }\cS\mbox{ is true}\Big).
	\end{matrix}	
\end{equation}
Inspired by possible world semantics, we propose
\begin{defn}
	\label{d:mgs}
	The \textbf{meaning} of output $y$ by mechanism $\fm$ is
	\begin{equation}
		\begin{matrix}
			p_{unif}(X) & \rightarrow & p_\fm(X|y)\\
			\Big(\mbox{possible inputs}\Big) & \rightarrow & 
			\Big(\mbox{inputs that cause }y\Big).
		\end{matrix}
	\end{equation}
	For a deterministic function this reduces to $\X\supset f^{-1}(y)$.
\end{defn}

Grounding meanings in mechanisms yields four advantages over the possible worlds approach. First, it replaces the difficult to define notion of a possible world with the concrete set of inputs the mechanism is physically capable of receiving. Second, in possible world semantics the work of determining whether or not a sentence is true is performed somewhat mysteriously offstage, whereas the meaning of a measurement is determined via Bayes' rule. Third, the approach generalizes to probabilistic mechanisms. Finally, we can compute the effective information generated by a measurement, whereas there is no way to quantify the information content of a sentence in possible world semantics.

\subsection{Risk}
\label{s:pragmatics}

The second, pragmatic notion of meaning characterizes usefulness. We consider a special case, well studied in statistical learning theory, where usefulness relates to predictions \cite{vapnik:98}.

Let $\Sigma(\X,\Y)=\{\sigma:\X\rightarrow \Y\}$ be the set of all functions (deterministic mechanisms) mapping $\X$ to $\Y=\{-1,+1\}$. We will often write $\Sigma$ for short. Suppose there is a random variable $X$ taking values in $\X$ with unknown distribution $P$ and an unknown mechanism $\sigma^*\in\Sigma$, the \emph{supervisor}, who assigns labels to elements of $\X$. 

\begin{defn}
	The \textbf{risk} quantifies how well mechanism $f$ approximates an unknown or partially known mechanism $\sigma^*$:
	\begin{equation}
		\label{e:risk}
		\risk(f) = \sum_{x\in \X} \bI\big[f(x)\neq \sigma^*(x)\big]\cdot p(x).
	\end{equation}
	It is the probability that $f$ and $\sigma^*$ disagree on elements of $\X$.
\end{defn}
Unfortunately, the risk cannot be computed since $P$ and $\sigma^*$ are unknown. 
\begin{defn}
	Given a finite sample $\data=(x_1,\ldots,x_l)\in\X^l$ with labels $\slabel=\sigma^*\data=(y_1,\ldots,y_l)\in\Y^l$, the \textbf{empirical risk} of $f:\X\rightarrow\Y$
	\begin{equation}
		\label{e:emp-risk}
		\risk(f,\data,\slabel)=\frac{1}{l}\sum_{i=1}^l\bI\big[f(x_i)\neq y_i\big]
	\end{equation}	
	 is the fraction of the data $\data$ on which $f$ and $\sigma^*$ disagree.
\end{defn}
The empirical risk provides a computable approximation to the (true) risk.

\begin{rem}
	\label{r:finite}
	Note that in this paper, sets $\X$ and $\Y$ are both finite. Similarly, the training data $\data\in \X^l$ and labels $\slabel\in\Y^l$ also live in finite sets.
\end{rem}

\section{Statistical learning theory}
\label{s:slt}

Suppose we wish to predict the unknown supervisor $\sigma^*$ based on its behavior on labeled data $(\data,\slabel)$. A simple way to find a mechanism in repertoire $\cF\subset \Sigma(\X,\Y)$ that approximates $\sigma^*$ well is to minimize the empirical risk. 
\begin{defn}
	Given repertoire $\cF\subset\Sigma$ and unlabeled data $\data\in\X^l$, define \textbf{learning algorithm}
	\begin{equation}
		\label{e:algol}
		\cA_{\cF, \data}:\Sigma  \rightarrow \cF:\sigma \mapsto \arg\min_{f\in\cF}\risk(f,\data,\sigma\data)
	\end{equation}
	which finds the mechanism in $\cF$ that minimizes empirical risk.
\end{defn}

Learning algorithm $\cA_{\cF,\data}$ finds the mechanism in $\cF$ that appears, based on the empirical risk, to best approximate $\sigma^*$. Empirical risk stays constant or decreases as $\cF$ is enlarged, suggesting that the larger the repertoire the better.

This is not true in general since minimizing risk -- and \emph{not} empirical risk -- is the goal. There is a tradeoff: increasing the size of $\cF$ leads to overfitting the data which can increase risk even as empirical risk is reduced.

The tendency of a repertoire to overfit data depends on its size or capacity. We recall two measures of capacity that are used to bound risk: empirical VC-entropy \cite{vapnik:82} and empirical Rademacher complexity \cite{koltchinskii:01}.

\begin{defn}
	Given unlabeled data $\data\in\X^l$ and repertoire $\cF\subset\Sigma$ let
	\begin{equation}
		q_\data:\cF\rightarrow \bR^l:f\mapsto\Big(f(x_1),\ldots,f(x_l)\Big).
		\label{e:vc-ent}
	\end{equation}
	The empirical \textbf{VC-entropy}\footnote{VC-entropy is the \emph{expectation} of empirical VC-entropy \cite{vapnik:98}. Also, note the standard definition of VC-entropy uses $\log_e$ rather than $\log_2$.} of $\cF$ on $\data$ is $\cV(\cF,\data):=\log_2 |q_\data(\cF)|$, where $|q_\data(\cF)|$ is the number of distinct points in the image of $q_\data$.
	
	The empirical \textbf{Rademacher complexity} of $\cF$ on $\data$ is
	\begin{equation}
		\radem(\cF,\data) = \frac{1}{|\Sigma|}
		\sum_{\sigma\in\Sigma}\left[\sup_{f\in\cF}\frac{1}{l}\sum_{i=1}^l\sigma(x_i)\cdot f(x_i)\right].
		\label{e:rademacher}
	\end{equation}
\end{defn}
VC-entropy ``counts'' how many labelings of $\data$ the classifiers in $\cF$ fit perfectly. Rademacher complexity is a weighted count of how many labelings of $\data$ functions in $\cF$ fit well.

The following theorems are shown in \cite{boucheron:00} and \cite{bousquet:04} respectively:
\begin{thm}[empirical VC-entropy bound]\label{t:vc}\eod
	With probability $1-\delta$, the expected risk is bounded by
	\begin{equation}
		\risk(f)   \leq \risk(f,\data,\slabel)
		  + c_1\sqrt{\frac{\cV(\cF,\data)}{l}}
		 + c_2\sqrt{\frac{1-\log_2\delta}{l}}
	\label{e:bd-vc}
	\end{equation}
	for all $f\in \cF$, where the constants are $c_1=\sqrt{\frac{6}{\log_2 e}}$ and $c_2=\sqrt{\frac{1}{\log_2 e}}$. 
\end{thm}

\begin{thm}[empirical Rademacher bound]\label{t:rademacher}\eod	
	For all $\delta>0$, with probability at least $1-\delta$,
	\begin{equation}
		\risk(f)\leq \risk(f,\data,\slabel)+ \radem(\cF,\data)+c_3\sqrt{\frac{1-\log_2\delta}{l}},
		\label{e:bd-rad}
	\end{equation}
	for all $f\in \cF$, where $c_3=\sqrt{\frac{2}{\log_2 e}}$.
\end{thm}
The tradeoff between empirical risk and capacity is visible in the first two terms on the right-hand sides of the bounds.

The left-hand sides of Eqs~\eqref{e:bd-vc} and \eqref{e:bd-rad} cannot be computed since $P$ and $\sigma^*$ are unknown. Remarkably, the right-hand sides depend only on mechanism $f$ chosen from repertoire $\cF$, labeled data $(\data,\slabel)$ and desired confidence $\delta$. The theorems assume data is drawn \emph{i.i.d.} according to $P$ and labeled according to $\sigma^*$; it make no assumptions about the distribution $P$ on $\X$ or supervisor $\sigma^*$, except that they are \emph{fixed}.

\section{Falsification}
\label{s:lim}

This section reformulates the results from statistical learning theory to show how the past falsifications performed by a learning algorithm control future performance. We show that the empirical VC-entropies and Rademacher complexities admit interpretations as ``counting'' (in senses made precise below) the number of hypotheses falsified by a particular measurement  performed when learning. 

We start by introducing a special mechanism, the min-risk, which is used implicitly in learning algorithm $\cA_{\cF,\data}$. As we will see, the structure of the measurements performed by the min-risk determine the capacity of the learning algorithm.
\begin{defn}
	\label{d:error}
	Given repertoire $\cF\subset\Sigma$ and unlabeled data $\data\in\X^l$, define the \textbf{min-risk} as the minimum of the empirical risk on $\cF$:
	\begin{equation}
		\label{e:error}
		\errmap_{\cF,\data}:\Sigma \rightarrow \bR:\sigma  \mapsto\min_{f\in\cF} \risk(f,\data,\sigma\data).
	\end{equation}
\end{defn}

\begin{figure}[thpb]
	\centering
	\includegraphics[scale=.8]{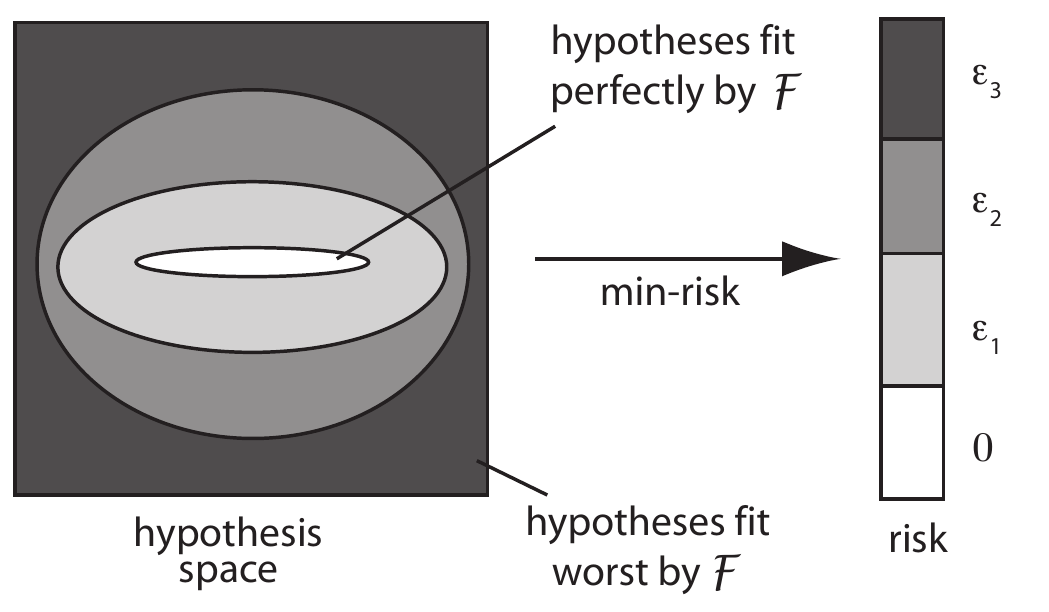}
	\caption{\textbf{The structure of the measurement performed by the min-risk.}
	\footnotesize{
	The min-risk categorizes potential hypothesis in $\Sigma$ according to how well they are fit by mechanisms in theory $\cF$.
	}}
	\label{f:mrisk}
\end{figure}

The min-risk is a mechanism mapping supervisors $\sigma$ in $\Sigma$ to the empirical risk of their best approximations $\cA_{\cF,\data}(\sigma)$ in $\cF$, see Fig. \ref{f:mrisk}. Note that inputs to the min-risk are themselves mechanisms. 

We suggestively interpret the setup as follows. Suppose a scientist studies a universe where inputs in $\X$ appear according to distribution $P$, and are assigned labels in $\Y$ by unknown physical process $\sigma^*$. The \emph{hypothesis space} is $\Sigma(\X,\Y)$, the set of all possible (deterministic) physical processes that take $\X$ to $\Y$.

The scientist's goal is to learn to predict physical process $\sigma^*$, on the basis of a small sample of labeled data $(\data,\slabel)$. She has a \emph{theory}, repertoire $\cF$, and a method, $\cA_{\cF,\data}$, which she uses to fit some particular  $\hat{f}\in\cF$ given $\slabel$. 

The most important question for the scientist is: How reliable are predictions made by $\hat{f}$ on \emph{new} data? We will show that $\hat{f}$'s reliability depends on the measurements performed by the min-risk --  i.e. on the work done by the scientist when she applies method $\cA_{\cF,\data}$ to find $\hat{f}$.

\subsection{Empirical VC entropy}
\label{s:rvc}

Empirical VC-entropy is, essentially, the effective information generated by the min-risk when it outputs a perfect fit:
\begin{prop}[VC-entropy via effective information]\label{t:ei-vc}\eod
	Empirical VC entropy is 
	\begin{equation}
		\cV(\cF,\data) = l - ei\left(\errmap_{\cF,\data}, 0\right).
		\label{e:ei-vc}
	\end{equation}
\end{prop}

\noindent
Proof: 
Let $\X=\data\cup\data^c$ and $|\X|=m$. Then $\Sigma=\{\sigma:\data\rightarrow \Y\}\times \{\sigma:\data^c\rightarrow \Y\}$. By definition
\begin{equation*}
		ei\left(\errmap_{\cF,\data}, 0\right) = \log_2 |\Sigma| - \log_2| \errmap^{-1}_{\cF,\data}(0)|,
\end{equation*}
with $\log_2 |\Sigma|=m$. It remains to show that $| \errmap^{-1}_{\cF,\data}(0)|=2^{m-l}\cdot |q_\data(\cF)|$. Points in the image of $q_\data$ correspond to labelings $\sigma$ of the data by functions in $\cF$. Thus, $|q_\data(\cF)|$ counts distinct labelings of $\data$ that $\cF$ fits perfectly. These occur with multiplicity $2^{m-l}$ in the pre-image by the product decomposition of $\Sigma$ above.
$\blacksquare$

We interpret the result as follows.  Suppose the scientist applies theory $\cF$ to explain her labeled data and perfectly fits function $\hat{f}=\cA_{\cF,\data}(\sigma^*)$ with risk $\epsilon=0$. 

By Definition~\ref{d:mgs}, the meaning of her work is $\Sigma \supset \errmap_{\cF,\data}^{-1}(0)$: the set of mechanisms that her theory $\cF$ fits perfectly. The effective information generated by her work is
\begin{equation}
	\begin{matrix}
		ei(\errmap_{\cF,\data},0) & = & \log_2\left|\Sigma\right| & -  & \log_2|\errmap_{\cF,\data}^{-1}(0)|\\
		& = & \Big(\mbox{total no. of hypotheses}\Big) & - & \Big(\mbox{no. that theory fits}\Big) \\
		& = & \Big(\mbox{no. of hypotheses falsified}\Big),
	\end{matrix}	
	\label{e:falsify}
\end{equation}
where hypotheses are counted in bits (after logarithming). A theory is informative if it rules out many potential hypotheses \cite{popper:59}.

\emph{The number of hypotheses the scientist falsifies when using theory $\cF$ to fit $\hat{f}$ has implications for its future performance}:

\begin{cor}[information-theoretic empirical VC bound]\label{t:ei-vcb}\eod
	With probability $1-\delta$, the risk of predictor $\hat{f}=\cA_{\cF,\data}(\sigma^*)$ outputted by learning algorithm $\cA_\cF$ is bounded by
	\begin{equation}
		\risk(f)  \leq \risk(f,\data,\slabel) + c_1\sqrt{1-\frac{ei(\errmap_{\cF,\data},0)}{l}}
		+ c_2\sqrt{\frac{1-\log_2 \delta}{l}}.
	\label{e:ei-vcb}
	\end{equation}	
\end{cor}

\noindent
Proof: By Theorem~\ref{t:vc} and Proposition~\ref{t:ei-vc}.
$\blacksquare$

The corollary states that minimizing empirical risk embeds expectations about the future into predictors. So long as the corollary's assumptions hold, future performance by $\hat{f}$ is controlled by: {(\rm i)} the output of the min-risk, i.e. the fraction $\epsilon$ of the data that $\hat{f}$ fits; {(\rm ii)} the effective information generated by the min-risk, i.e. the number (in bits) of hypotheses the learning algorithm falsifies if it fits perfectly; and {(\rm iii)} a confidence term. The only assumption made by the corollary is that $P$ and $\sigma^*$ are \emph{fixed}.

\begin{rem}
	The theorem provides no guarantees on the future performance of a theory that ``explains everything'', i.e. $\cF=\Sigma$, no matter how well it fits the data. This follows since effective information is zero when $\cF=\Sigma$, and so the second term on the right-hand side of Eq.~\eqref{e:ei-vcb} is $c_1\approx 2$. 
\end{rem}

Reformulating the above result in terms of code lengths suggests a connection between VC-theory and minimum message length (MML), see \cite{wallace:68} and \S6.6 of \cite{dowe:10}. Recall that, given probability distribution $p(X)$, the message length of event $x$ in an optimal binary code is $\text{len}(x):=-\log_2 p(x)$. 

\begin{cor}[VC-entropy controls code length of true hypothesis]\label{t:mml}\eod	
	Denote the min-risk by $\fm=\risk_{\cF,\data}$. The length of the true hypothesis $\hat{\sigma}$ in the optimal code for the actual repertoire specified by the min-risk,  $p_\fm(\Sigma|\epsilon=0)$, is 
	\begin{equation*}
		\text{len}(\hat{\sigma}) = \cV(\cF,\data) + \big(|\X|-|\data|\big).
	\end{equation*}
\end{cor}

\noindent
Proof: By Proposition \ref{t:ei-vc} we have $-\log_2 p_\fm(\hat{\sigma}|\epsilon=0) = \log_2 |\risk_{\cF,\data}^{-1}(0)|$.
$\blacksquare$

The length of the message describing the true hypothesis in the actual repertoire's optimal code is the empirical VC-entropy plus a term, $(|\X|-|\data|)=(m-l)$, that decreases as the amount of training data increases. The shorter the message, the better the predictor's expected performance (for fixed empirical risk).

\subsection{Empirical Rademacher complexity}
\label{s:rrad}

VC-entropy only considers hypotheses that theory $\cF$ fits perfectly. Rademacher complexity is an alternate capacity measure that considers the distribution of risk across the entire hypothesis space. This section explains Rademacher complexity via an analogy with Solomonoff probability \cite{solomonoff:64, wallace:99}. 

We first recall Solomonoff's definition. Given universal Turing machine $T$, define (unnormalized) \textbf{Solomonoff probability}
\begin{equation}
	\label{e:solomonoff}
	p_T(s) := \sum_{\{i|T(i)= s\bullet\}} 2^{-\text{len}(i)},
\end{equation}
where the sum is over strings\footnote{A technical point is that no proper prefix of $i$ should output $s$.} $i$ that cause $T$ to output $s$ as a prefix, and $\text{len}(i)$ is the length of $i$. We adapt Eq.~\eqref{e:solomonoff} by replacing Turing machine $T$ with min-risk $\risk_{\cF,\data}:\Sigma\rightarrow \bR$. 

\begin{defn}
	Equipping hypothesis space with the uniform distribution $p_{unif}(\Sigma)$, all hypotheses have length $\text{len}(\sigma)=|\X|=\log_2|\Sigma|$ in the optimal code. Set the \textbf{Rademacher distribution} for the min-risk $\fm=\risk_{\cF,\data}$ as
	\begin{equation}
		\label{e:rad_dist}
		p_\fm(\epsilon) :=
		\sum_{\left\{\sigma|R_{\cF,\data}(\sigma)=\epsilon\right\}}2^{-\text{len}(\sigma)}=\left\{\begin{matrix}
		\frac{\big|\errmap_{\cF,\data}^{-1}(\epsilon)\big|}{|\Sigma|} & 
		\mbox{ if }\epsilon\in \errmap_{\cF,\data}(\Sigma)\\ \\
		0 & \mbox{ else.}
	\end{matrix}\right.
	\end{equation}
\end{defn}

The Rademacher distribution is constructed following Solomonoff's approach after substituting the min-risk as a ``special-purpose Turing machine'' that only accepts hypotheses in finite set $\Sigma$ as inputs. It tracks the fraction of hypotheses in $\Sigma$ that yield risk $\epsilon$.

The Rademacher distribution arises naturally as the denominator when using Bayes' rule to compute the actual repertoire $p_\fm(\Sigma|\epsilon)$:
\begin{equation*}
	p_\fm(\sigma|\epsilon)=\frac{p_\fm(\epsilon|\sigma)}{p_\fm(\epsilon)}\cdot p_{unif}(\sigma),
	\,\,\,\text{ where }p_\fm(\epsilon|\sigma)=\left\{\begin{matrix}
	1 & 
	\mbox{ if }\errmap_{\cF,\data}(\sigma)=\epsilon\\ \\
	0 & \mbox{ else.}
\end{matrix}\right.
\end{equation*}

\begin{prop}[Rademacher complexity via min-risk]\label{t:mr-rad}\eod
	\begin{equation}
		\radem(\cF,\data)=1-2\cdot \bE\big[\epsilon\,\big|\,p_\fm(\epsilon)\big].
	\end{equation}
\end{prop}

\noindent
Proof: We refer to $\bE\big[\epsilon\,\big|\,p_\fm(\epsilon)\big]$ as the expected min-risk. From Eq.~\eqref{e:rademacher},
\begin{equation*}
	\radem(\cF,\data) = \frac{1}{|\Sigma|}
	\sum_{\sigma\in\Sigma}
	\left[\sup_{f\in\cF}\frac{1}{l}\sum_{i=1}^l\sigma(x_i)\cdot f(x_i)\right].
\end{equation*}
Observe that $\frac{1}{l}\sum_{i=1}^l\sigma(x_i)\cdot f(x_i) = 1-2\risk(f,\data,\sigma)$. It follows that $\sup_{f\in\cF}\frac{1}{l}\sum_{i=1}^l\sigma(x_i)\cdot f(x_i)=1-2\errmap_{\cF,\data}(\sigma)$, which implies 
\begin{equation*}
	\radem(\cF,\data) = 1-2\sum_{\sigma\in \Sigma}
	\frac{\errmap_{\cF,\data}(\sigma)}{|\Sigma|} 
	= 1-2\sum_{\epsilon} \epsilon\cdot \frac{\big|\errmap_{\cF,\data}^{-1}(\epsilon)\big|}{|\Sigma|}.
	\,\,\blacksquare
	\label{e:rad-int}
\end{equation*}
\vspace{1mm}

Rademacher complexity is low if the expected min-risk is high. The expected min-risk admits an interesting interpretation. For any hypothesis $\sigma\in\errmap_{\cF,\data}^{-1}(\epsilon)$ the classifier $\hat{f}_\sigma:=\cA_{\cF,\data}(\sigma)\in\cF$ outputted by the learning algorithm  yields incorrect answers on fraction $\epsilon=\frac{1}{l}\sum_{i=1}^l\bI\big[\hat{f}_\sigma(x_i)\neq \sigma(x_i)\big]$ of the data. It follows that 
\begin{equation*}
	\begin{matrix}
	\sum_\epsilon p_\fm(\epsilon)\cdot \epsilon & = & \sum_\epsilon &
	\frac{\big|\errmap_{\cF,\data}^{-1}(\epsilon)\big|}{|\Sigma|}
	&\cdot&
	\frac{1}{l}\sum_l \bI\big[\hat{f}_\sigma(x_i)\neq\sigma(x_i)\big]
	\\
	& = & \sum_\epsilon &\Big(\mbox{fraction of hypotheses falsified}\Big)&\cdot &\Big(\mbox{on fraction }\epsilon\mbox{ of the data}\Big).
	\end{matrix}
\end{equation*}

A bold theory $\cF$ is one for which $\bE[\epsilon|p_\fm(\epsilon)]$ is high, meaning that its predictors (the classifiers it tries to fit to data) are sufficiently narrow that it would falsify most hypotheses on most of the data. 

\emph{When a bold theory happens to fit labeled data well, it is guaranteed to perform well in future:}
\begin{cor}[information-theoretic empirical Rademacher bound]\label{t:ei-radb}\eod
	With probability $1-\delta$, the risk of predictor $\hat{f}=\cA_\cF(\data,\slabel)$ outputted by learning machine $\cA_\cF$ is bounded by
	\begin{equation}
			\risk(f) \leq \risk(f,\data,\slabel) 
			+ \left[1-2\sum_{\epsilon} \epsilon\cdot 2^{-ei(\errmap_{\cF,\data},\epsilon)}\right]
			+ c_3\sqrt{\frac{1-\log_2\delta}{l}}
	\end{equation}	
\end{cor}
Proof: By Proposition~\ref{t:mr-rad} and definition of effective information we have
\begin{equation*}
	\radem(\cF,\data)=1-2\sum_{\epsilon} \epsilon\cdot \frac{\big|\errmap_{\cF,\data}^{-1}(\epsilon)\big|}{|\Sigma|}=1-2\sum_{\epsilon} \frac{\epsilon}{2^{ei(\errmap_{\cF,\data},\epsilon)}}.
\end{equation*}
The result follows by Theorem~\ref{t:rademacher}.
$\blacksquare$

Rademacher complexity is low if the min-risk's sharp measurements (high $ei$) are accurate (low $\epsilon$), and conversely. Analogously to Corollary~\ref{t:ei-vcb}, the Rademacher bound implies the future performance of a classifier depends on: {(\rm i}) the fraction $\epsilon$ of the data that $\hat{f}$ fits; {(\rm ii)} the weighted (by the fraction $\epsilon$ of data that falsifies them) sum of the fraction of hypotheses falsified; and {(\rm iii}) a confidence term. Once again, the only assumption is that $P$ and $\sigma^*$ are \emph{fixed}.

\section{Discussion}
\label{s:discuss}

Learning according to algorithm $\cA_{\cF,\data}$ entails computing the min-risk, which classifies hypotheses about $\data$ according to how well they are approximated by predictors in repertoire $\cF$. Repertoires that rule out many hypotheses when they fit labeled data $(\data,\slabel)$ generate more effective information than repertoires that ``approximate everything''. As a consequence, when and if an informative repertoire fits labeled data well, Corollary~\ref{t:ei-vcb} implies we can be confident in future predictions on unseen data. 

A pleasing consequence of reformulating empirical VC-entropy and empirical Rademacher complexity in terms of falsifying hypotheses is that it directly connects Popper's intuition about falsifiable theories to statistical learning theory, thereby providing a rigorous justification for the former.

Our motivation for reformulating learning theory information-theoretically arises from a desire to better understand the role of information in biology. Although Shannon information has been heavily and successfully applied to biological questions, it has been argued that it does not fully capture what biologists mean by information since it is not semantic. For example, Maynard Smith states that ``In biology, the statement that A carries information about B implies that A has the form it does because it carries that information'' \cite{maynardsmith:00}. Shannon information was invented to study communication across prespecified channels, and lacks any semantic content. Maynard Smith therefore argues that a different notion of information is needed to understand in what sense evolution and development embed information into an organism.

It may be fruitful to apply statistical learning theory to models of development. One possible approach is to consider analogs of repertoire $\cF$. For example, $\cF$ may correspond to the repertoire of possible adult forms a zygote could develop into. The particular adult form chosen, $\hat{f}\in\cF$, depends on the historical interactions $(\data,\slabel)$ between the organism and its environment, assuming these can be suitably formalized. The information generated by the organism's development would then have implications for its future interactions with its environment. More speculatively, a similar tactic could be applied to quantify the information embedded in populations by inheritance and natural selection.

{
\footnotesize\small
\bibliographystyle{splncs03}

}
\end{document}